# Neural networks trained with WiFi traces to predict airport passenger behavior

Federico Orsini[1]

Massimiliano Gastaldi[1,2]

Luca Mantecchini[3]

Riccardo Rossi[1]

[1]Department of Civil, Environmental and Architectural Engineering, University of Padova, Italy
[2]Department of General Psychology, University of Padova, Italy
[3]DICAM, School of Engineering and Architecture, University of Bologna, Italy

# Neural networks trained with WiFi traces to predict airport passenger behavior

Federico Orsini
ICEA Department
University of Padova
Padova, Italy
federico.orsini@unipd.it

Massimiliano Gastaldi
ICEA Department
Department of General Psychology
University of Padova
Padova, Italy
massimiliano.gastaldi@unipd.it

Luca Mantecchini
DICAM, School of Engineering and Architecture
University of Bologna
Bologna, Italy
luca.mantecchini@unibo.it

Riccardo Rossi
ICEA Department
University of Padova
Padova, Italy
riccardo.rossi@unipd.it

*Abstract*— The use of neural networks to predict airport passenger activity choices inside the terminal is presented in this paper. Three network architectures are proposed: Feedforward Neural Networks (FNN), Long Short-Term Memory (LSTM) networks, and a combination of the two. Inputs to these models are both static (passenger and trip characteristics) and dynamic (real-time passenger tracking). A real-world case study exemplifies the application of these models, using anonymous WiFi traces collected at Bologna Airport to train the networks. The performance of the models were evaluated according to the misclassification rate of passenger activity choices. In the LSTM approach, two different multi-step forecasting strategies are tested. According to our findings, the direct LSTM approach provides better results than the FNN, especially when the prediction horizon is relatively short (20 minutes or less).

## I. INTRODUCTION

With an ever rising number of flights in the world (average growth rate on annual basis at 3.8% by 2034, as reported in IATA, 2015) an efficient airport passenger management is essential, as it can have a great impact on both customer satisfaction and the airport financial performance ([2], [3]). It is reported that a pleasant airport experience encourages spending and influences future travel plans ([4], [5]). Retail concession business is often an important element in the overall financial portfolio of companies operating in airports [6]. In some cases non-aviation revenues account for as much as 53% of total revenues [7]. Recently some projects focusing on airport passengers have been carried out: Airport of the Future Project (AFP) focuses on safety, security and efficiency, and has been applied in Australian airports; Proactive Passenger Flow Management for Airports with an Advanced Forecasting System (AERFOR) has developed a tool for airport management, by predicting passenger flows inside the terminal; and DORA is an extended project to reduce passengers' travel time, providing them with real-time route information from an origin (e.g., their home) to the aircraft.

Understanding and modelling pedestrian behavior is essential for a good airport passenger management. The growing interest of airport operators for passenger flow management and business intelligence strategies has produced several theoretical studies and experimental applications to passengers tracking. Several methods and technologies have been applied to collect data and to simulate passenger flows, in order to measure queue dynamics, transit times and waiting times at processing points. Although a number of works have been published about this variety of topics (e.g., [11]–[13]), they mainly focus on walking behavior, crowd dynamics and passenger operations, instead of activity choice.

In the literature there are some examples of activity-based approaches to model airport passenger's behavior. Liu et al. [14], presented an activity-based travel demand model, focusing on activity scheduling of airport passengers, based on revealed and stated preference survey data. The model had a nested structure which divided possible activities at the airport into three phases. They examined the behavior of passengers with differing socio-demographic and flight characteristics. Kalakou and Moura [15] focused on passengers' activity choice before security control. A multinomial logit model was estimated to predict passenger's choices. Data were collected with a revealed preference survey. In the micro-simulation model of Ma [16] passengers were defined as agents with some initial basic and advanced traits, and were divided into several groups in terms of route-choice preferences. The study used empirical data to validate the model; the use of surveys and video cameras was suggested. Recently, Jiang and Ren [17] presented a study focusing on passenger's behavior in unexpected situations, such as flight delays.

In the era of big data, more and more machine learning techniques are applied in many engineering areas, and the air transportation field makes no exception. Recently, Chen and Li [18] developed a wavelet neural network to make short-term predictions of passenger flow in the terminal; Fatemi Ghomi and Forghani [19] compared neural networks and Box-Jenkins method in order to forecast airline passenger demand; whereas Xia et al. [20] focused on the prediction of air route passenger flows. However, the use of these techniques has not yet been applied to model passenger activity choice inside airport terminals.

In this work we propose the use of neural networks to model passenger behavior inside airport terminals and, more specifically, to predict the passenger's activity sequence. In the next chapter the neural networks architecture is presented in general terms; in the third chapter a real-world case study is presented and shows the application of the proposed techniques; the fourth chapter concludes the paper with some remarks and indications for future developments.

## II. METHODOLOGY

### A. General framework and objectives

This paper is part of a wider project, whose aim is to develop a decision support system for providing information and suggestions to airport users with a smartphone application. This system will be able to improve passengers' experience, by reducing time spent queueing and waiting, and to raise airport revenues, by increasing the time passengers spend in discretionary activities. A detailed description of the whole system can be found in [21], where originally a discrete choice approach to model passengers' behavior, similar to that of Danalet [22], was theorized.

In this work a different approach is followed and the passenger's behavior is modelled with neural networks. The main reason behind this choice was the possibility to work with real-time information on passenger movements, thanks to the network architecture, and the low computational power that the model requires.

In the model presented in this paper airport passenger activities are divided into different categories. The time horizon is defined as a span of time before the flight scheduled departure, and is discretized into several time units. The activity choices made by passengers are defined according to activity type and time unit, i.e., passengers choose a certain activity type at a certain time unit.

### B. Neural networks architecture

Three neural networks architectures are proposed in this paper.

1. Feed-forward architecture (FNN);
2. Long-short term memory architecture (LSTM)
3. A combination between the two.

In the next paragraphs we will walk through these approaches under an application-oriented point of view. Formal aspects of these models can be widely found in the literature (e.g., [23], [24]).

#### 1) FNN architecture

The first approach proposed is a very simple feed-forward architecture. FNN is the simplest type of artificial neural network: information moves in only one direction, forward, from the input nodes, through the hidden nodes and to the output nodes, without cycles or loops [25].

In this paper's case, the only inputs are the general passenger information. It is possible to divide these inputs into two different categories:

1. Passenger personal characteristics, such as: gender, age, income, education, frequent flyer or not, etc.
2. Passenger trip characteristics, such as: time of arrival at the airport, destination of the flight, flight carrier, etc.

While personal characteristics mainly contain socio-demographic information that may be difficult to obtain, some of the trip characteristic can be deduced even from anonymous traces.

Under the assumption that each activity performed at a given time unit is independent from the other activities performed in previous time units, it is possible to train independently as many FNN for as many time units of the model. For each time unit, the output of the corresponding FNN is the activity chosen at that time unit by a passenger with given characteristics (see Figure 1).

#### 2) LSTM architecture

In this approach we relax the disputable assumption that activity choice is independent from the activities performed during previous time units. On the contrary, the only inputs are the activity performed during the previous time units.

Long short-term memory (LSTM) networks are a type of recurrent neural network (RNN) that can learn long-term dependencies between time steps of sequence data [24]. The inputs to these kind of networks are time series. LSTMs can be used for sequence classification [26], sequence-to-sequence classification or regression [27], and, as in the case of this paper, for time series forecasting. There is increasing interest in transportation-related LSTM applications ([28], [29]).

Like RNNs, LSTMs use the output from a previous step as an input for the next step. Nodes perform calculations using the inputs and returning an output value. This output is then used along with the next element as the input for the next step, and so on. In an LSTM network, nodes also have an internal state, which is used as a working memory space, where information can be stored and retrieved over many time steps. Output values are determined by input values, previous outputs, and the internal state; results of nodes calculations are used to both provide an output value and to update the internal state. Like in RNNs, LSTM nodes have parameters that determine how the inputs are used in the calculations; in addition to this, they have gates, which control how much the saved state information is used as an input to the calculations. Similarly, there are gates to control how much of the current information is saved to the internal state, and how much the output is determined by the current calculation and by the saved information. LSTM nodes are more complicated than RNN nodes, but this helps them at learning the complex interdependencies in sequences of data (see Figure 1).

Intuitively, the situation presented in this work can be looked at in analogy to the case of text generation, where the task is to predict the next character in a stream of text. The "time series" in that case is, in fact, a sequence of characters; the output (i.e., the next character), depends both on short-term memory (e.g., previous characters in the same word) and long-term memory (e.g., the context) [30].

In the airport passenger case, a code is assigned to each activity type, and an activity sequence becomes a time series composed by a sequence of these codes, one for each time unit. The LSTM network task is to predict the next activity in the sequence, which may depend not just on the activity performed immediately before, but also on activities performed further before.

In many applications, it is useful to make predictions not just on the next time-step, but on several future time steps. In order to do this, two main multi-step forecasting strategies exist: the *direct* and the *recursive* ([31], [32]).

In the *recursive strategy* the base LSTM network is used multiple times, with the prediction for the prior time step being used as an input for making a prediction on the following time step. The main drawback of this strategy is that, since predictions are used in place of observations, prediction errors tend to accumulate, and the performance can quickly degrade as the prediction time horizon increases.

In the *direct strategy*, several independent LSTM networks are trained, for each prediction horizon: therefore, a network is trained to predict which activity will be performed two time units ahead, another to predict what will happen three time units ahead, and so on. In this case, error propagation is avoided, although the forecasting performance still decreases, as the prediction horizon expands. Another drawback is that larger computational time is required, as the number of networks to train increases.

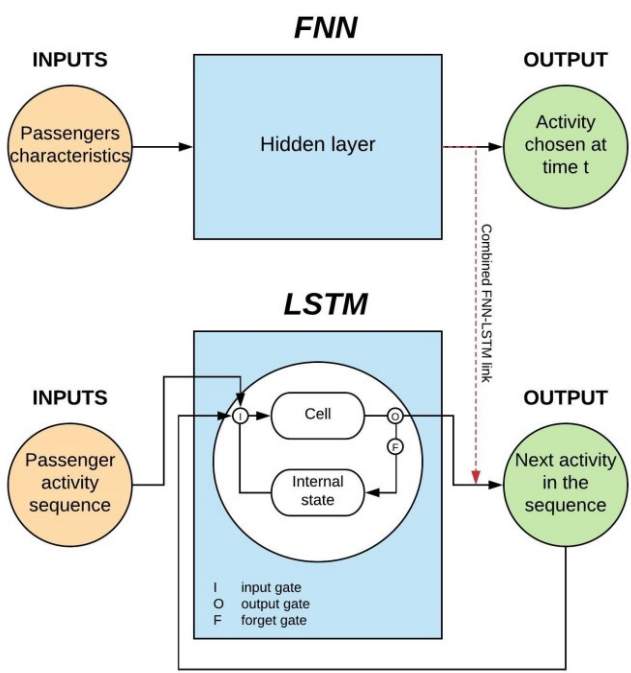

Fig. 2. Neural networks architecture scheme

*3) Combined architecture*

The classic implementation LSTM, despite being able to include the complex interdependencies between the activity choices within the sequence, is unable to incorporate "static" features, such as the passenger's characteristics.

The need to have a model which includes both dynamic and static inputs can be found in clinical applications. For example, Esteban et al. [33] needed to predict, for a given patient, whether a transplantation endpoint would occur within six or twelve months after the visit, based on a time series of medical data collected in the past (e.g., laboratory tests, medication prescribed, etc.), and also on static information (e.g., age, gender, blood type, etc). They developed a combined network, processing the static information on an independent FNN and the dynamic information with a LSTM, before concatenating the hidden states of both networks and providing this information to the output layer.

The same approach can be applied in the airport passenger case, by merging the two architectures presented in the previous paragraphs before the output layer. In this way, it is possible to incorporate in a single neural network both the activity sequence and the passenger characteristics (see Figure 1).

III. CASE STUDY

*A. Data acquisition*

Bologna Airport is a major international airport in Italy; with over 8 million of carried passengers in 2017, it is ranked 8th in the country, while it is 6th in the ranking based on total number of commercial movements (67,088) and 5th for cargo carried (41,986 Tons) [34].

The airport offers a free WiFi service. Passengers who connect to the network can be tracked and are identified with their MAC address. In this case study WiFi traces were collected during 2 months, from December 2017 to January 2018. The airport was divided into 32 areas; each of them was associated to a certain activity type and contained a number of WiFi access points. Data were pre-processed by airport staff, and contained for each MAC address the time instant at which the passenger entered a certain area and the time instant at which the passenger left the same area. In this way it was possible to reconstruct the activity sequence for each MAC address; moreover, it was possible to associate a number of passengers to their flight.

A total number of 192,925 unique MAC address were tracked during the 2-months period of interest. However, many of the traces had to be discarded for multiple reasons: (i) some passengers were not tracked to the boarding gate, so their MAC address could not be linked to a specific flight; (ii) some passengers were not tracked continuously and their activity sequences contained "data gaps"; (iii) some passengers started to be tracked only after security control.

After filtering the data, 5805 high quality activity sequences remained; 70% of these were used in the model calibration, while the remaining 30% were kept for the model validation.

In this activity choice model 6 different activity types were considered, defined in the following way:

- Mandatory: when the passenger is detected queueing or being serviced at check-in desks, security control or customs control;

- Eating when the passenger is detected within an area containing bars or restaurants;

- Shopping when the passenger is detected within an area containing shops;

- Waiting: when the passenger detected in a waiting area near the gates;

- Other: when the passenger detected in other areas;
- Not-at-the-airport: from the start of the time horizon until the first detected activity.

Time horizon and time unit were set respectively at 180 and 5 minutes. In the case in which passengers performed more than one activity within the same unit, the activity performed for the longest time was selected as the activity type associated to that time unit.

### B. Analysis of results

#### 1) FNN

As mentioned in Chapter II, the FNN architecture is composed by several networks, one for each time unit and independent from the others, taking as inputs passenger characteristics.

The data collected and described in the previous paragraph were anonymous. Unfortunately, this is a relevant limit for the model, since none of the passenger personal characteristics were available. However, it was possible to deduce some trip characteristic: arrival time at the airport (time of the day); earliness (time before departure); destination; carrier; smartphone brand.

Therefore, the inputs of the FNNs were the following:

- Arrival time (hour of the day, normalized);
- Earliness (minutes before scheduled departure time, normalized);
- Destination (dummy: 1 short range, 0 medium range, according to Eurocontrol definition [35]);
- Carrier (dummy: 1 traditional carrier, 0 low-cost carrier);
- Smartphone brand (dummy: 1 Apple, 0 Android or other).

All the FNNs had a 3-layer structure, with one input layer, one hidden layer and an output layer. The input layer contained as many nodes as the input variables (therefore 5), and the output layer as many nodes as the activity classes (therefore 6). To determine the hidden layer size, several FNNs architecture were trained and tested, progressively increasing the number of hidden nodes. A hidden layer size of 6 produced the best performance, in terms of misclassified activities in the test dataset.

Figure 2 presents the results obtained with the FNN approach, in terms of misclassified activity types. Performance is particularly good at the beginning and at the end of the time horizon, when there is less variation in the possible range of the activity performed. For example, 3 hours before the flight almost 90% have not yet entered the airport (therefore they are performing the "not-at-the – airport" activity), while 10 minutes before departure time more than 95% of the passengers are in the waiting areas or already on the plane. The highest misclassification value (about 45%) occurs 50 minutes before the departure flight, when the spread of passengers across the activity types is very high.

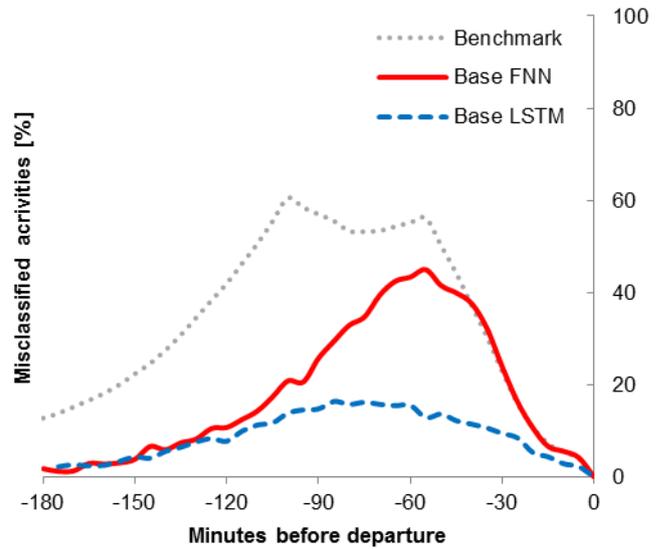

Fig. 2. Number of misclassified activities across the time horizon for FNN, LSTM and benchmark models.

To evaluate the importance of the various inputs, others FNNs architecture were implemented, removing one different input variable each time. To assess the impact of the single variables, the average misclassification rate in the "critical period" (i.e., between 100 and 30 minutes before the flight departure time) was calculated (see Table I). As a benchmark, Table also contains the average misclassification rate of a FNN in which the inputs were random numbers.

TABLE I.  EVALUATION OF INPUT VARIABLES IMPORTANCE

| Model | Mean misclassification rate during critical period |
|---|---|
| Base model | 32.3% |
| Without arrival time | 33.1% |
| Without earliness | 49.6% |
| Without destination | 32.4% |
| Without carrier | 32.5% |
| Without smartphone brand | 32.9% |
| Benchmark with random numbers | 52.0% |

It is evident that the most important variable is the earliness; without it, results are comparable with the random benchmark; on the contrary, the removal of any other variable produces results only marginally worse than the base model. Indeed, passenger earliness is recognized as a key factor for airport planning and management, and methodologies to estimate arrival rate functions have been recently developed [36].

It is worth noting that the destinations that can be reached from Bologna are mainly European cities with a flight distance lower than 1500 km (therefore short-range flights according to Eurocontrol definition [35]), and few destinations farther but still within 4000 km (medium-range). A bigger heterogeneity in destination distances may produce significant differences.

The model predictions were evaluated also against another trivial benchmark: a model in which all passengers were assigned, at each time step, to the most performed activity type. With respect to this benchmark, Figure 2 shows a significant improvement in the results, which however, in relative terms, tends to decrease along the time horizon. In the last 30 minutes, in fact, the two models produce very similar results. This can be interpreted in this way: the FNN output is mainly influenced by passenger earliness, whose impact on activity choice decrease progressively; in the last half-hour before departure most of passengers will be either at a waiting area or inside the airplane, regardless of how early they arrived at the airport.

*2) LSTM*

In the LSTM architecture, a single network is trained using as inputs passenger's activity sequences; at each time unit of the input sequence, the LSTM network learns to predict the activity performed during the next time unit, i.e., the model is able to predict what happens one time unit ahead, given the sequence of activities performed before.

The LSTM hidden layer had a size of 200, and the network was trained with the stochastic gradient descent with momentum (SGDM) optimizer. Figure 2 shows the misclassification ratio, for each time unit. A big improvement can be observed with respect to the FNN results, especially during what it was previously defined as the "critical period" (100 to 30 minutes before flight departure). In that period the mean misclassification rate dropped from the 32.3% of the base FNN model to only 13.9%.

The results are very much satisfactory for applications in which a reliable short-term prediction is needed. However, an airport manager may be interested not only in what will happen in the next five minutes, but also in the medium term. Therefore, it would be interesting to produce a multi-step forecasting of the activity sequences. As mentioned in Chapter II, two main multi-step forecasting strategies exist: the recursive strategy and the direct strategy.

Figure 3 shows the results obtained implementing the recursive strategy. Each curve represents the misclassification rate for a different prediction horizon, from 5 minutes (1 time unit ahead, i.e., the base LSTM network) to 30 minutes (6 time units ahead). As a benchmark, also the FNN base model results are shown. From the figure it is possible to observe how the performance of the networks degrades, as the prediction horizon increases. The misclassification rate is higher than in the FNN model in the first part of the sequence, because in the FNN model, one of the inputs is passenger earliness: this allows the model to generate very good predictions for the activity type "not at the airport", which is prominent in the early part of the time horizon. In the critical period multi-step LSTM networks over perform the FNN benchmark when the prediction horizon is less than 20 minutes. The issue of error propagation is particularly evident toward the end of the time horizon, and it escalates as the prediction horizon increases: for the 30 minute curve, the final activity is misclassified in about 34% of the cases, while in the LSTM base model this rate is only 2.3%.

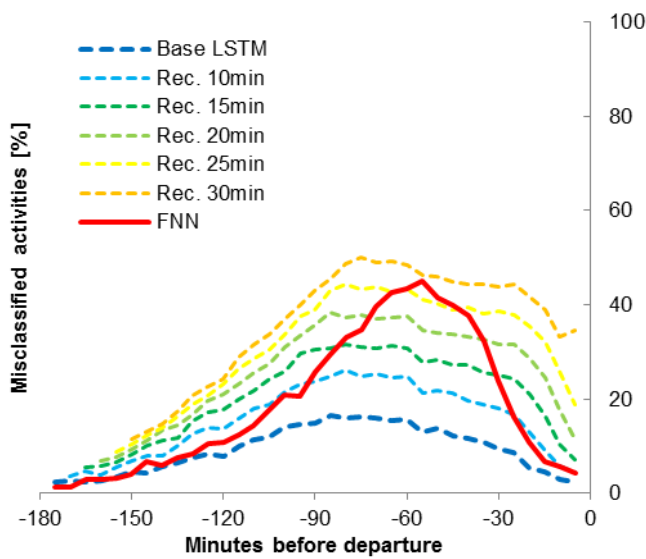

Fig. 3. Recursive strategy. Number of misclassified activities across the time horizon, for different prediction horizons.

The direct strategy produces slightly better results during the critical period (see Figure 4 and Table II): an improvement of 1.8% for the 10 minute horizon, which increases up to 5.8% for the 20 minute horizon. In this case the 20 minute curve still performs better than the FNN benchmark. More importantly, it is possible to notice that the issue of error propagation disappears, and all networks are more or less equally able to correctly predict the final part of the activity sequences.

The combined FNN-LSTM architecture is not presented in this case study. As it was shown in the previous paragraph, the only significant input in the FNN model is the passenger earliness. This trip characteristic is actually implicitly considered also in the base LSTM model, since the passenger arrival at the airport is identified in the sequence by a switch from the "not at the airport" activity to another activity type. Therefore, combining the two networks would not provide any additional information.

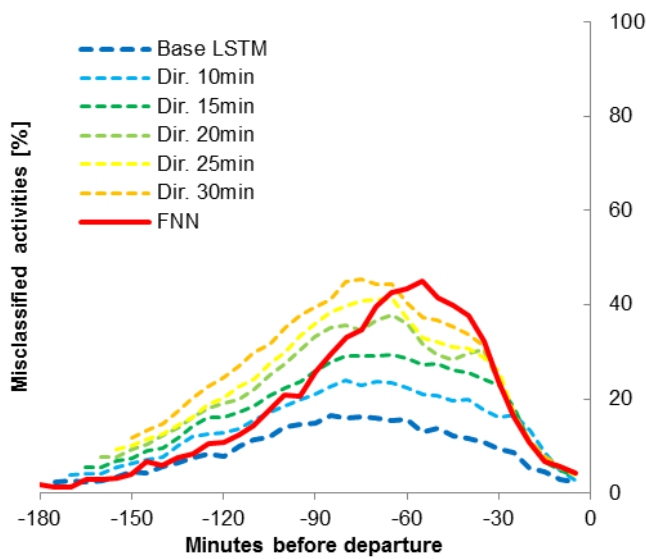

Fig. 4. Direct strategy. Number of misclassified activities across the time horizon, for different prediction horizons.

TABLE II. MISCLASSIFICATION RATES DURING CRITICAL PERIOD FOR DIFFERENT FORECASTING STRATEGIES AND PREDICTION HORIZONS

| Prediction horizon | Recursive strategy [%] | Direct strategy [%] | Difference [%] |
|---|---|---|---|
| 5 minutes | 13.92 | 13.92 | 0.0 |
| 10 minutes | 22.40 | 20.60 | -1.80 |
| 15 minutes | 28.37 | 25.93 | -2.44 |
| 20 minutes | 34.13 | 31.44 | -2.68 |
| 25 minutes | 39.21 | 34.09 | -5.12 |
| 30 minutes | 43.85 | 38.03 | -5.82 |

*C. Real-world application*

As mentioned before, this paper is part of a wider project, with the final objective of optimizing passengers time at the airport, by providing targeted suggestions able to reduce queueing time and increasing the time devoted to discretionary activities. In that framework, it is crucial to predict the activity sequence of passengers, and in particular the time at which they decide to move toward security/customs controls. Prediction of airport passenger's behavior are essential to develop these kind of recommendation systems, and the outputs from the models presented in this work can also be used in other frameworks (e.g., [37]).

When a passenger arrives at the airport and opens the airport smartphone application, passenger and trip characteristics will be collected by filling in a quick form. This information will be used to predict the passenger activity sequence using the FNN architecture. The smartphone application will monitor the movements of the passenger and the activities carried out; the predicted activity sequence will then be regularly updated with the combined FNN-LSTM network, improving the accuracy of the predictions in the short/medium term (5 to 20 minute prediction horizon).

This will allow the system to know when a given passenger wishes to reach security control, and, eventually, to suggest the passenger a different time when to do so. More details on the suggestion generation can be found in [21].

IV. CONCLUSION

In this paper a machine-learning-based approach for predicting airport passenger's behavior was proposed. Feedforward Neural Networks (FNN) and Long Short-Term Memory (LSTM) networks were trained using real-world WiFi traces.

The FNN architecture used as inputs static passenger and trip characteristics, while the LSTM architecture information on activities performed before. A combined FNN-LSTM was also proposed, to exploit both static and dynamic information.

A real-world case study was presented, in order to show an example of practical application. In terms of misclassified activities, the performance of the FNN model was very good in the initial part of the passenger's activity sequences, before worsening in the critical period (100 to 30 minutes before the flight departure). The LSTM network provided much more reliable predictions in the short-term, whereas increasing the prediction horizon, the performance became comparable with that of the FNN model. Two different approaches were applied for the multi-step forecast, with the direct strategy giving, in general terms, better results than the recursive.

There are still some limits that will be tackled in future research.

The data used to train the neural networks of the case study were not collected with this kind of application in mind. The use of completely anonymous traces limited the power of the FNN approach; in the future, traces linked to passenger personal characteristics may significantly improve the performance of the model. In addition to this, if more input variables are found to have a significant impact on the prediction reliability of the FNN model, then the combined architecture of FNN-LSTM will be tested.

In the LSTM approach, only two basic forecasting strategies were tested. Future research will involve testing other more advanced approaches, such as DirRec or multiple-output strategies [32].

The models presented in this work are specifically parametrized to deal with departing airport passengers; however, after proper adaptation, this methodology can be applied in other transportation contexts.

ACKNOWLEDGEMENTS

The authors would like to thank Bologna Airport for providing access to the WiFi traces, and Neotecnica Srl for the technical support. This work was supported by the University of Padova and Regione Veneto (project: "Un sistema di supporto all'Utenza per la SCelta delle attIvità in un Terminal Aeroportuale (USCITA)". Grant number: FSE 2105-97-2216-2016).